\documentclass[11pt]{article}

\usepackage{EMNLP2022}

\usepackage[utf8]{inputenc} 
\usepackage[T1]{fontenc}    
\usepackage{hyperref}       
\usepackage{url}            
\usepackage{booktabs}       
\usepackage{amsfonts}       
\usepackage{nicefrac}       
\usepackage{microtype}      
\usepackage{xcolor}         

\usepackage{graphicx}
\usepackage{amsmath}
\usepackage{algorithm}
\usepackage[noend]{algpseudocode}
\usepackage{wrapfig}

\title{AutoWS: Automated Weak Supervision Framework for Text Classification}

\author{%
  Abhinav Bohra \\
  Amazon.com LLC\\
  \texttt{abohra@amazon.com} \\
  \And
  Huy Nguyen \\
  Amazon.com LLC \\
  \texttt{nguynnq@amazon.com} \\
  \And
  Devashish Khatwani \\
  Amazon.com LLC \\
  \texttt{khatwad@amazon.com} \\
}

\begin{document}

\maketitle

\begin{abstract}
Creating large, good quality labeled data has become one of the major bottlenecks for developing machine learning applications. Multiple techniques have been developed to either decrease the dependence of labeled data (zero/few-shot learning, weak supervision) or to improve the efficiency of labeling process (active learning). Among those, Weak Supervision has been shown to reduce labeling costs by employing hand crafted labeling functions designed by domain experts. We propose \emph{AutoWS} -- a novel framework for increasing the efficiency of weak supervision process while decreasing the dependency on domain experts. Our method requires a small set of labeled examples per label class and automatically creates a set of labeling functions to assign noisy labels to numerous unlabeled data. Noisy labels can then be aggregated into probabilistic labels used by a downstream discriminative classifier. Our framework is fully automatic and requires no hyper-parameter specification by users. We compare our approach with different state-of-the-art work on weak supervision and noisy training. Experimental results show that our method outperforms competitive baselines. 
\end{abstract}

\section{Introduction}

Text classification is among the most popular Natural Language Processing (NLP) tasks, and has important applications in real-world, e.g., product categorization. The advent of Deep Learning has brought the state-of-the-art to a wide variety of text classification tasks \citep{minaee_deep_2021}. However, this comes with a price: deep learning based models usually require large amounts of annotated training data to achieve superior performance. In many scenarios, manual data annotation is expensive in terms of cost and effort, especially when subject matter experts (SME) must involve and/or classification tasks are scaled to millions of instances with hundreds to thousands of classes. To reduce the manual annotation effort, machine learning research have explored weak supervision approaches, i.e., possibilities to build prediction models using limited, noisy or imprecise labeled data. Weak supervision concerns special conditions of supervised learning in which annotated training data may be incomplete, inexact or inaccurate \citep{zhou_brief_2018}.
In this paper, we study incomplete supervision, i.e., data has ground truth labels but is too small to train a performant model, and focus on semi-supervised learning techniques to leverage numerous amount of in-domain unlabeled data to improve prediction performance.

Prior studies have covered different methods to assign noisy training labels to unlabeled data including crowdsourcing \citep{dalvi_aggregating_2013,joglekar_comprehensive_2015,yuen_survey_2011,zhang_spectral_2016}, distant supervision \citep{hoffmann_knowledge-based_2011,mintz_distant_2009,smirnova_relation_2018,takamatsu_reducing_2012}, heuristic rules \citep{awasthi_learning_2020,ratner_snorkel_2017,ratner_data_2016,varma_inferring_2017}. Pre-trained language models gain much attention recently because they can be fine-tuned with little annotated data thanks to their great generalization capability \citep{perez_true_2021,radford_language_2019}. The above weak supervision sources are termed \emph{labeling functions} (LF), which may vary in terms of error rate, coverage and probably generate conflict labels \citep{zhang_survey_2022,zhang_wrench_2021}. Researchers have developed \emph{label models} that aggregate output of labeling functions to generate confidence-weighted or probabilistic labels, which are consequently used to train an end model (e.g., final text classifier).

In this study, we propose \emph{AutoWS} -- an automated end-to-end weak supervision framework for text classification. AutoWS provides a wide range of machine-learning based labeling functions ranging from statistical models to transformer-based language models, and different label models.
Aiming to provide users a fully automated framework, AutoWS implements a simple yet effective evaluation process even it is provided just a small labeled dataset. With our proposed evaluation process, only top labeling functions and the best label model are selected to produce final labels to unlabeled data.
Contributions of our study are followings:

\begin{itemize}
  \item AutoWS is a fully automated framework so that users neither need to provide any labeling heuristics/functions nor tune hyper-parameters.
  \item Our experiments cover a wide variety of data domains including news, users' request, product titles, and emphasizes datasets with a large number of classes. While weak supervision has been studied for many years, its application on many-class classification tasks was not comprehensively evaluated.
  \item With a capability of selecting top labeling functions and best label models, AutoWS outperforms prior studies on benchmark data.
\end{itemize}

\section{Related Work}

In past few years, research in weak-supervision explored automatic labeling function generation to remove the reliance on subject matter experts. Snuba system used three learning algorithms, i.e., Decision Tree, Logistic Regressions, and K-Nearest Neighbor, to learn multiple classifiers by varying feature sets \citep{varma_snuba_2018}. \textsc{TaLLoR}, GLaRA systems based on a set of simple logical rules (i.e., predicates) to induce compound rules for using as labeling functions in a named entity tagging task \citep{li_weakly_2021,zhao_glara_2021}. Despite differences among approaches, those studies aimed to maintain a large and diverse pool of labeling functions so that label aggregation models can work more effectively. In our study, we create a pool of 15 labeling functions which are learned automatically from labeled data. Therefore, our approach does not require human-created seed functions or probing feature subsets as in prior studies.

Pre-training then fine-tuning framework shows great effectiveness in many classification tasks, especially weak supervision \citep{radford_language_2019}.
Notable work in this direction includes ELMo~\citep{peters_deep_2018}, BERT~\citep{devlin_bert_2019}, ULMFiT~\citep{howard_universal_2018} and VAMPIRE~\citep{gururangan_variational_2019}. Self-training further enhances this paradigm by incorporating both labeled and unlabeled data in an unified iterative fine-tuning process.
\citep{clark_semi-supervised_2018,miyato_adversarial_2016,xie_unsupervised_2020}. In our study, instead of improving fine-tuning methods, we make use of large pre-trained language models to create strong base classifiers as labeling functions. As a result, our approach empowers the labeling function pool by increasing its prediction power rather than using a large number of weak predictors (e.g. rules, heuristics).

To systematically resolve conflicts and overlapping among noisy labeling functions, advanced label aggregation techniques have been proposed to replace majority voting.
Representative work from weak supervision include Data Programming \citep{ratner_data_2016}, MeTaL \citep{ratner_training_2019} and FlyingSquid \citep{fu_fast_2020}. In our study, we utilize and compare different label aggregation methods (i.e. \emph{Label Model}) to achieve the best final classification model. An advantage of AutoWS is a simple yet effective evaluation procedure which allows us select the best method among candidates.

\section{AutoWS Architecture}

\begin{figure*}
  \centering
  \includegraphics[width=\textwidth]{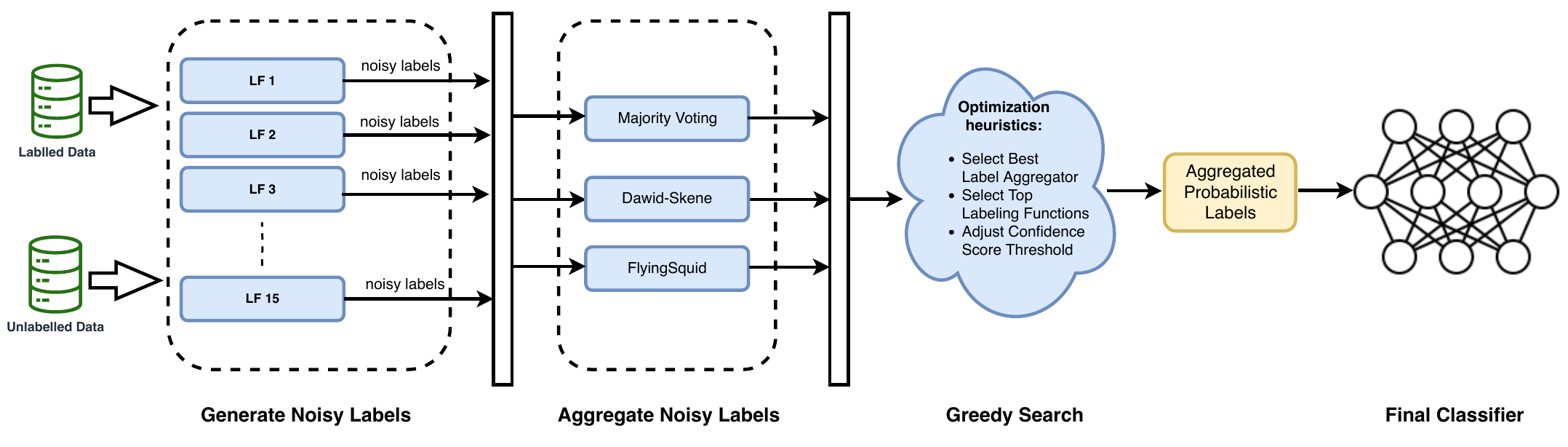}
  \caption{AutoWS architecture}
  \label{fig:arch}
\end{figure*}

AutoWS is a multi-module pipeline that has four main components: (1) learn \emph{Labeling Functions} from labeled datasets, (2) generate probabilistic labels using \emph{Label Models}, (3) greedy search the best label model and top labeling functions, and (4) learn final classification model. AutoWS architecture is illustrated in Figure~\ref{fig:arch}.

\subsection{Problem Setup}

We are given a set $\mathbf{X} = \{x_1, x_2, ..., x_L\}$ of $L$ labeled samples with corresponding ground-truth labels $\mathbf{Y} = \{y_1, y_2, ..., y_L\}$, where $y_i$ is label of sample $x_i$ with $i = 1..L$, $y_i \in [0, C - 1]$, $C$ is number of label classes. Let set $\mathbf{U} = \{u_1, u_2, ..., u_N \}$ of $n$ unlabeled samples. For convenience, we also use $Y_x$ to indicate label value of data point $x$. Weak supervision setup assumes $\mathbf{X}$ and $\mathbf{U}$ are drawn from the same distribution and $L \ll N$. Let $\mathit{F} = \{f_1, f_2, ..., f_M\}$ be a set of $M$ labeling functions. Let a label assignment to $u \in \mathbf{U}$ be $f_i(u) \in [-1, C - 1]$ where $f_i(u) = -1$ indicates $f_i$ abstains from giving a label to $u$. For an input $x$, let $\bar{x}$ be a feature vector by a feature extraction function, e.g., Tf-Idf vectorizer.

\subsection{Labeling Functions}

AutoWS makes use of total 15 labeling functions which is a blend of heuristics, statistical machine learning and deep learning.

\paragraph{Feature-based Models}
To reduce feature engineering, we only consider word-vectorization methods to transform each raw input into a feature vector. Three word-vectorizers are utilized:

\begin{itemize}
  \item Word-count vectorizer\footnote{Scikit-learn's CountVectorizer}: this transforms each input text into a word count vector. In current implementation, we only extract unigrams given small labeled data.
  \item Tf-Idf vectorizer\footnote{Scikit-learn's TfidfVectorizer}: this converts input text into term-frequency inverse-document-frequency vector. Similarly, we only extract unigrams.
  \item Sentence embedding: A pre-trained sentence transformer model is executed to generate embeddings of input text \citep{reimers_sentence-bert_2019}. We use pre-trained model \emph{paraphrase-MiniLM-L6-v2}\footnote{\url{www.sbert.net/docs/pretrained\_models.html}}
\end{itemize}

Once inputs are transformed into feature space, we use three supervised learning algorithms to train classification models.
For each algorithm, we implement an Bayesian Optimizer to fine-tune following learning algorithm's hyper-parameters through a K-fold cross validation on labeled data.

\begin{itemize}
  \item Adaboost: number of trees, maximum tree depth, learning rate, minimum number of samples in leaf node, 	and number of features when searching best split during branching.
  \item Random Forest: number of trees, and maximum tree depth.
  \item Support Vector Machine: regularization parameter $C$ and kernel coefficient $\gamma$.
\end{itemize}

We utilize Scikit-learn library to train 9 labeling models of this group \citep{pedregosa_scikit-learn_2011}.

\paragraph{Similarity-based Models}
Labeling functions in this group will assign label to an input when it is found similar to a labeled sample. We build three heuristics using above feature extraction: word count, Tf-IDF, and sentence embeddings. For similarity-based labeling, each unlabeled sample is assigned the label of most similar labeled sample. Sample similarity $\mathtt{sim}(u, x)$ is the cosine similarity between two feature vectors $\bar{u}, \bar{x}$. Let $x^* = \max_{x \in  \mathbf{X}}(\mathtt{sim}(u, x))$, we have $Y_u = Y_{x^*}$.


When there are multiple $x^*$ that are equally most similar to $u$ but have different labels, we assign $u$ the label that is most popular in $\mathbf{X}$. After the above process, each unlabeled sample $u$ is associated with a label $Y_u$ and a similarity score $\mathtt{sim}_u = \mathtt{sim}(u, x^*)$. 
We observed that in some cases, $\mathtt{sim}_u$ is low and similarity-based labeling may not reliable. To handle such cases, we rank $u \in \mathbf{U}$ by their similarity score $\mathtt{sim}_u$ from high to low and assign label -1 to samples below 10th quantile.

\paragraph{Transformer-based Models}
This group consists of three transformer-based models: BERT \citep{devlin_bert_2019}, ELECTRA \citep{clark_electra_2020}, and DeBERTa \citep{he_deberta_2021}.
AutoWS uses pre-trained BERT (base-uncased version), ELECTRA (base) and DeBERTa (base) and fine-tune the networks on labeled data. While prior work experimented with weak, simple labeling heuristics, we think strong labeling functions can contribute better to yield higher quality probabilistic labels for unlabeled data. We use AutoGluon's text predictor to train these models \citep{shi_multimodal_2021}.

\subsubsection{Scoring Labeling Functions}
\label{sec:lf_score}

We hypothesize that not all labeling functions help a label model generate good quality probabilistic labels. Thus, we implement an optimization module that selects top labeling functions through an iterative process (see Section~\ref{sec:optimization}). An exhaustive search over all combination of 15 labeling functions is too computationally expensive, so we propose a greedy search by ranking labeling functions by prediction score. We split labeled set $\mathbf{X}$ in to $\mathbf{X}_\mathrm{train}$ and $\mathbf{X}_\mathrm{dev}$ of 80\% and 20\% of samples respectively. We use $\mathbf{X}_\mathrm{train}$ to train labeling functions and report prediction accuracy on $\mathbf{X}_\mathrm{dev}$. Once labeling functions are evaluated, they are retrained with complete dataset $\mathbf{X}$. Labeling functions' accuracy scores are provided to optimization module (Section~\ref{sec:optimization})

\subsection{Label Models}

Unlabeled data is passed through newly trained labeling functions to generate noisy labels (each data point is assigned multiple noisy labels). Then noisy labels are aggregated to generate final labels for unlabeled data. Current implementation of AutoWS includes three label aggregators: Majority Voting heuristic, Dawid-Skene model \citep{dawid_maximum_1979} and FlyingSquid model \citep{fu_fast_2020}. We implement unweighted majority voting so that it simply counts how many LF vote for a label without considering label prior or LF evaluation score. Abstaining LF does not vote.


Dawid-Skene model \citep{dawid_maximum_1979} assumes an unobserved or latent ground truth label $Y_u$ of sample $u$ and the probabilities that LF $f_i$ provides correct or incorrect label $f_i(u)$ are only dependent on $Y_u$
FlyingSquid method models the joint probability distribution of hidden true labels and observed LF outputs as a Binary Ising model when simplifying the classification task into binary version \citep{fu_fast_2020}. Each LF $f$ is augmented by two binary observed variables $f^0$ and $f^1$ which are dependent to latent ground truth label. A Triplet Method is used to solve the optimization problem through a closed form set of equations so no learning is needed.

\subsection{Final Classification Model}

Given a set of labeling functions (LF), a label model and a numerical threshold $t$, we first run LF on unlabeled data $\mathbf{U}$ to generate label matrix $\mathcal{L}$ where $\mathcal{L}_{ij}$ keeps LF output $f_i(u_j)$. Label matrix is fed into label model to generate probabilistic labels for every $u \in \mathbf{U}$. We discard samples $u$ whose labels have aggregated probabilistic score less than $t$,\footnote{We utilize output score returned by label model implementation in WRENCH} and let the remaining set $\mathbf{U}' = \{u: p(u) \ge t\}$. Finally, to train the final classification model, we fine-tune DeBERTa with the union set $\mathbf{X} \cup \mathbf{U}'$.

\subsubsection{Optimization Module}
\label{sec:optimization}

With multiple labeling functions and label models, we propose a greedy search algorithm to find the best label model and top labeling functions. As mentioned above, for each configuration of a LF set, a label model and a threshold $t$, we obtain a dataset $\mathbf{U}'$ with weak labels. We evaluate the prediction power of those weak labels by fine-tuning DeBERTa using $\mathbf{U}'$ and record prediction performance on $\mathbf{X}$.\footnote{This process is different than training the final classifier in that it uses $\mathbf{X}$ as test data rather than training data}. We perform the following search heuristic:

\begin{enumerate}
  \item With label model set $\mathbf{A}$ and probability threshold set $\mathbf{T}$, we generate $|\mathbf{A}| \times |\mathbf{T}|$ configurations in which each configuration includes all LF, a label model $A \in \mathbf{A}$ and a threshold $t \in \mathbf{T}$. By evaluating these configurations, we obtain the best label model $A^*$, and threshold $t^*$.
  \item With a set of quantiles $\mathbf{Q}$ we generate $|\mathbf{Q}|$ configurations in which each has top $q$ LF ($q \in \mathbf{Q}$), label model $A^*$, and threshold $t^*$. Evaluating these configurations give us the best $q^*$ to select top LF.
\end{enumerate}

Provided the best configuration of top $q^*$ LF, label model $A^*$ and threshold $t^*$, we generate probabilistic labels for unlabeled data and train final classification model.

\section{Data and Baseline Models}

\begin{table}
  \caption{Dataset statistics}
  \label{data}
  \centering
  \begin{tabular}{lrrr}
    \toprule
    Dataset & \#Classes & Train/Dev & Test \\
    \midrule
    AGNews & 4 & 120,000 & 7,600  \\
    Yahoo! & 10 & 1,400,000 & 60,000  \\
    Retail & 21 & 41,586 & 4,642 \\
    \bottomrule
  \end{tabular}
\end{table}

In order to evaluate AutoWS and compare with different baseline models, we experiment with three many-class public datasets as shown in Table~\ref{data}. First two datasets are \emph{AGNews} -- news articles \citep{zhang_character-level_2015} and \emph{Yahoo!} -- user questions/answers \citep{chang_importance_2008}.
We use the data split by MixText and compare our results with this study \citep{chen_mixtext_2020}.
MixText achieved the best results among prior semi-supervised learning methods on the data by adapting the novel Mixup idea. MixText relies on transformer-based models to encode textual inputs into hidden vectors so that they can interpolate two inputs to create a new training instance. Furthermore MixText uses predictions on unlabeled samples and their corresponding back-translations to assign labels to the unlabeled set.


Our third dataset is Retail \citep{elayanithottathil_retail_2021} comprising of 46,228 product items
We compare AutoWS with the best noisy-label model reported in \citep{nguyen_robust_2022}. In the study, the authors applied noise-resistance training to address a challenge that training data has incorrect labels. Weak labels generated by AutoWS are considered noisy. So this comparison gives us insight of applying noise-resistance training in weak supervision.



\section{Experimental Results}

\subsection{AGNews and Yahoo! Datasets}

Following MixText \citep{chen_mixtext_2020}, we sub-sample 200 inputs per class from original training data to create labeled dataset $\mathbf{X}$, and compile unlabeled data with 5,000 samples per class. Test data is kept intact (see Table~\ref{data}). Results are shown in Table~\ref{tab:result1}. As shown in row SOTA, MixText achieved accuracy 89.2 for AGNews and 71.3 for Yahoo! data. These were reported significantly higher than prior weak supervision methods, i.e., VAMPIRE \citep{gururangan_variational_2019} and UDA \citep{xie_unsupervised_2020}. Our AutoWS with optimization outperforms MixText by yielding accuracy 89.4 and 72.3 for AGNews and Yahoo! data respectively.\footnote{Our results are average over 3 runs with different random seeds and every run returns better performance than MixText.} 

To demonstrate the added-value by weak supervision components of AutoWS, we report performance of other settings: (1) use only labeled data to train the final classifier, (2) run AutoWS without optimization (i.e., output of all labeling functions are provided to Majority Voting). In the first setting, all DeBERTa, ELECTRA and BERT performed significantly worse than AutoWS. This result is consistent with prior studies, and further proves the essential impact of leveraging noisy label data to performance improvement. In the second setting, by feeding output of all labeling function outputs to Majority Voting, we observe significant accuracy decrease. This proves the critical role of optimization module in AutoWS.

\begin{table}
  \caption{Prediction performance (accuracy $\times$ 100\%) on AGNews, Yahoo! datasets.}
  \label{tab:result1}
  \centering
  \begin{tabular}{l|rr}
    \toprule
     & \multicolumn{1}{c}{AGNews} & \multicolumn{1}{c}{Yahoo!} \\
    \midrule
    Train samples per class & 200 & 200 \\
    Unlabeled data total & 20,000 & 50,000 \\
    \midrule
    SOTA & 89.2 & 71.3 \\
    DeBERTa & 85.7 & 67.4 \\
    ELECTRA & 81.2 & 60.9 \\
    BERT & 88.3 & 69.6 \\
    AutoWS (wo/ Opt) & 88.1 & 70.0 \\
    AutoWS (w/ Opt) & \textbf{89.4} & \textbf{72.3}\\
    \bottomrule
  \end{tabular}
\end{table}




\subsection{Retail Dataset}

Retail data was used in \citep{nguyen_robust_2022} to study impact of noisy label (i.e., item label is incorrect) to prediction performance. The prior study experimented with training data which was corrupted to have 20\% and 40\% samples have incorrect labels. The authors trained an LSTM-CNNs model on noisy data, and found that CoTeachPlus training method \citep{yu_how_2019} achieved the best performance improvement.
CoTeachPlus co-trains two models simultaneously but let each model only pass samples with small loss and prediction disagreement to the other model for training.

AutoWS generates weak labels for unlabeled dataset, thus the final classifier is trained on noisy data. Performance on Retail data is shown in Table~\ref{tab:result2}. In our analysis, using 100 labeled samples per class, AutoWS generates weak label training data which has noise rate 21\%. The final classifier obtain F1 score 68.7 on test data. With 200 labeled samples per class, the generated weak labels has noise rate 17\% and final accuracy is improved to 73.1. This shows that increasing labeled data size will lead to higher quality weak label data. However, AutoWS performance is significantly lower than CoTeachPlus even though our weak label data has lower noise rate than the prior study. This reveals the limit of regular training on noisy data. In future, we plan to apply noise-resistance training methods, e.g., CoTechPlus, to train the final classifier.

\begin{table}
  \caption{Prediction performance (test macro F1 $\times$ 100\%) on Retail data}
  \label{tab:result2}
  \centering
  \begin{tabular}{l|l|r}
    \toprule
    Condition & Model & F1 \\
    \midrule
    Clean label & LSTM-CNNs & 82.0 \\
    \midrule
    Noise rate 20\% & CoTeachPlus & 78.5 \\
    Noise rate 40\% & CoTeachPlus & 73.2 \\
    \midrule
    100 per class & AutoWS & 68.7 \\
    200 per class & AutoWS & 73.1 \\
    \bottomrule
  \end{tabular}
\end{table}

\subsection{Labeling Function Ranking}

Given a pool of 15 labeling functions, we conduct an analysis to reveal how labeling functions (LF) perform in different settings. For each dataset, we record LF scores (see Section~\ref{sec:lf_score}), and rank LF from high to low. We observe that deep-learning based LF are not always at top positions. For example, in AGNews data, BERT has the highest prediction score but the second and third places belong to Support Vector Machine (SVM) and Random Forest (RF) with sentence embeddings feature. DeBERTa and ELECTRA takes 4th and 7th positions respectively. More interestingly, in Banking77 data, we see SVM, Similarity-based Labeling and RF are top-3. Both BERT and DeBERTa show up in top-3 LF in Yahoo! and Product datasets. These findings support our decision to keep a diverse set of LF with presence of heuristics, statistical machine learning and deep-learning.
In order to make AutoWS more robust, we plan to pre-train transformer-based language models for different domains, e.g., product, dialogue, to use as LF and final classifier.

\section{Conclusions}

In this paper, we described AutoWS -- a fully automated weak supervision framework which allows users to build competitive text classifiers with a small amount of labeled data. AutoWS implements a diverse set of labeling functions and three different label aggregation models so that it can produce probabilistic labels for unlabeled data. We proposed a simple yet effective optimization heuristic to select top labeling function and the best label model. This makes AutoWS more robust than other weak supervision systems which require user to provide labeling heuristics, choose label model or tune hyper-parameters. Experiment results show that AutoWS outperform prior weak supervision models on benchmark data.

In future, we will extend AutoWS with domain-specific models, and more label aggregation methods. Furthermore, we will enhance optimization module and equip AutoWS with noise-resistance training solutions to address noisy labels generated by the system. Last but not least, we plan to open-source AutoWS to gain more attentions and interests from not only NLP/ML experts but also developers and researchers outside of the fields. Having AutoWS employed by end-users with little to zero knowledge of NLP/ML to solve real-world text classification applications with limited data is our ultimate goal.


%

\bibliography{AutoWS}
\bibliographystyle{acl_natbib}

\end{document}